\pdfoutput=1
\documentclass[11pt]{article}
\usepackage{acl}
\usepackage[tikz]{bclogo}
\usepackage{times}
\usepackage{latexsym}
\usepackage{multirow}
\usepackage[T1]{fontenc}
\usepackage[utf8]{inputenc}
\usepackage{lipsum}
\usepackage{microtype}

\usepackage[english]{babel}

\usepackage{amsmath}
\usepackage{amssymb}
\usepackage{amsfonts}
\usepackage{amstext}
\usepackage{amsthm}

\usepackage{caption}
\usepackage{subcaption}

\usepackage{array, makecell}
\usepackage{amsmath,amsthm}
\usepackage{multirow}
\usepackage{microtype}
\usepackage{booktabs}
\usepackage{lipsum}  
\usepackage{longtable}
\usepackage{enumitem}
\usepackage{subcaption}
\usepackage{adjustbox}
\usepackage{tikz}

\usepackage[noend]{algorithmic}
\usepackage{algorithm}

\usepackage{inconsolata}
\usepackage{listings}

\colorlet{punct}{red!60!black}
\definecolor{background}{HTML}{EEEEEE}
\definecolor{delim}{RGB}{20,105,176}
\colorlet{numb}{magenta!60!black}

\lstdefinelanguage{json}{
    basicstyle=\scriptsize\ttfamily,
    numbers=left,
    numberstyle=\scriptsize,
    stepnumber=1,
    numbersep=8pt,
    showstringspaces=false,
    breaklines=true,
    frame=lines,
    backgroundcolor=\color{background},
    literate=
     *{0}{{{\color{numb}0}}}{1}
      {1}{{{\color{numb}1}}}{1}
      {2}{{{\color{numb}2}}}{1}
      {3}{{{\color{numb}3}}}{1}
      {4}{{{\color{numb}4}}}{1}
      {5}{{{\color{numb}5}}}{1}
      {6}{{{\color{numb}6}}}{1}
      {7}{{{\color{numb}7}}}{1}
      {8}{{{\color{numb}8}}}{1}
      {9}{{{\color{numb}9}}}{1}
      {:}{{{\color{punct}{:}}}}{1}
      {,}{{{\color{punct}{,}}}}{1}
      {\{}{{{\color{delim}{\{}}}}{1}
      {\}}{{{\color{delim}{\}}}}}{1}
      {[}{{{\color{delim}{[}}}}{1}
      {]}{{{\color{delim}{]}}}}{1},
}

\DeclareMathOperator{\expect}{\mathbb{E}}

\newcommand{\benchmarkname}[1]{NLU Linguistic Phenomena}
\newcommand{\challengename}[1]{\textsc{Curriculum}}

\title{\vspace*{-0.5in}
{{\small \hfill \textit{NAACL 2022}} \\
\vspace*{.25in}} \textsc{Curriculum}: A Broad-Coverage Benchmark for \\ Linguistic Phenomena in Natural Language Understanding}

\author{Zeming Chen \quad Qiyue Gao \\
  Rose-Hulman Institute of Technology \\
  \{chenz16, gaoq\}@rose-hulman.edu
}

\definecolor{msftBlue}{RGB}{0,164,239}
\definecolor{msftGreen}{RGB}{127,186,0}
\definecolor{msftYello}{RGB}{255,185,0}
\definecolor{msftBlack}{RGB}{0,0,0}

\def\colorModel{hsb} 

\newcommand\ColCell[1]{
  \pgfmathparse{#1<50?1:0}  
    \ifnum\pgfmathresult=0\relax\color{white}\fi
    
    \pgfmathsetmacro\compA{#1/100}   
    \pgfmathsetmacro\compB{1-#1/100} 
    \pgfmathsetmacro\compC{1}        
    
  \edef\x{\noexpand\centering\noexpand\cellcolor[\colorModel]{\compA,\compB,\compC}}\x #1
  } 
\newcolumntype{E}{>{\collectcell\ColCell}m{0.4cm}<{\endcollectcell}}  

\date{}

\begin{document}
\maketitle
\begin{abstract}
In the age of large transformer language models, linguistic evaluation play an important role in diagnosing models' abilities and limitations on natural language understanding. However, current evaluation methods show some significant shortcomings. In particular, they do not provide insight into how well a language model captures distinct linguistic skills essential for language understanding and reasoning. Thus they fail to effectively map out the aspects of language understanding that remain challenging to existing models, which makes it hard to discover potential limitations in models and datasets. In this paper, we introduce \textsc{Curriculum} as a new format of NLI benchmark for evaluation of broad-coverage linguistic phenomena. \textsc{Curriculum} contains a collection of datasets that covers 36 types of major linguistic phenomena and an evaluation procedure for diagnosing how well a language model captures reasoning skills for distinct types of linguistic phenomena. We show that this linguistic-phenomena-driven benchmark can serve as an effective tool for diagnosing model behavior and verifying model learning quality. In addition, our experiments provide insight into the limitation of existing benchmark datasets and state-of-the-art models that may encourage future research on re-designing datasets, model architectures, and learning objectives. \footnote{Our code and data are publicly available at \url{https://github.com/eric11eca/curriculum-ling}}.
\end{abstract}

\section{Introduction}
With the rising power of pre-trained language models, large-scale benchmarks serve as an important factor driving the future progress of NLP. These benchmarks can provide a tool for analyzing the strengths and weaknesses of pre-trained language models. In recent years, many benchmarks \cite{wang2019glue, wang2020superglue, rajpurkar2018know} have been proposed that offer a diverse set of evaluation objectives. However, recent criticisms have been made that these benchmarks fail to serve as effective measures of progress in machine learning \cite{raji2021ai}. In particular, the task design does not formulate specific linguistic skills required for understanding. They lack the effectiveness in helping researchers understand how certain systems or models work and how they fail. Although many state-of-the-art language models have shown impressive performance on these common benchmarks, their performance degrades considerably on adversarial or out-of-distribution samples \cite{BrasSBZPSC20}. The performance drop shows that models may not be learning the required linguistic skills for solving the tasks of these benchmarks but exploit spurious dataset biases \cite{poliak-etal-2018-hypothesis}. Overall, the current benchmark format seems to be more like a contest than a tool that can explain how well a language model captures distinct linguistic skills essential to language understanding and reasoning.

\begin{figure}[t!]
    \centering
    \includegraphics[width=0.48\textwidth]{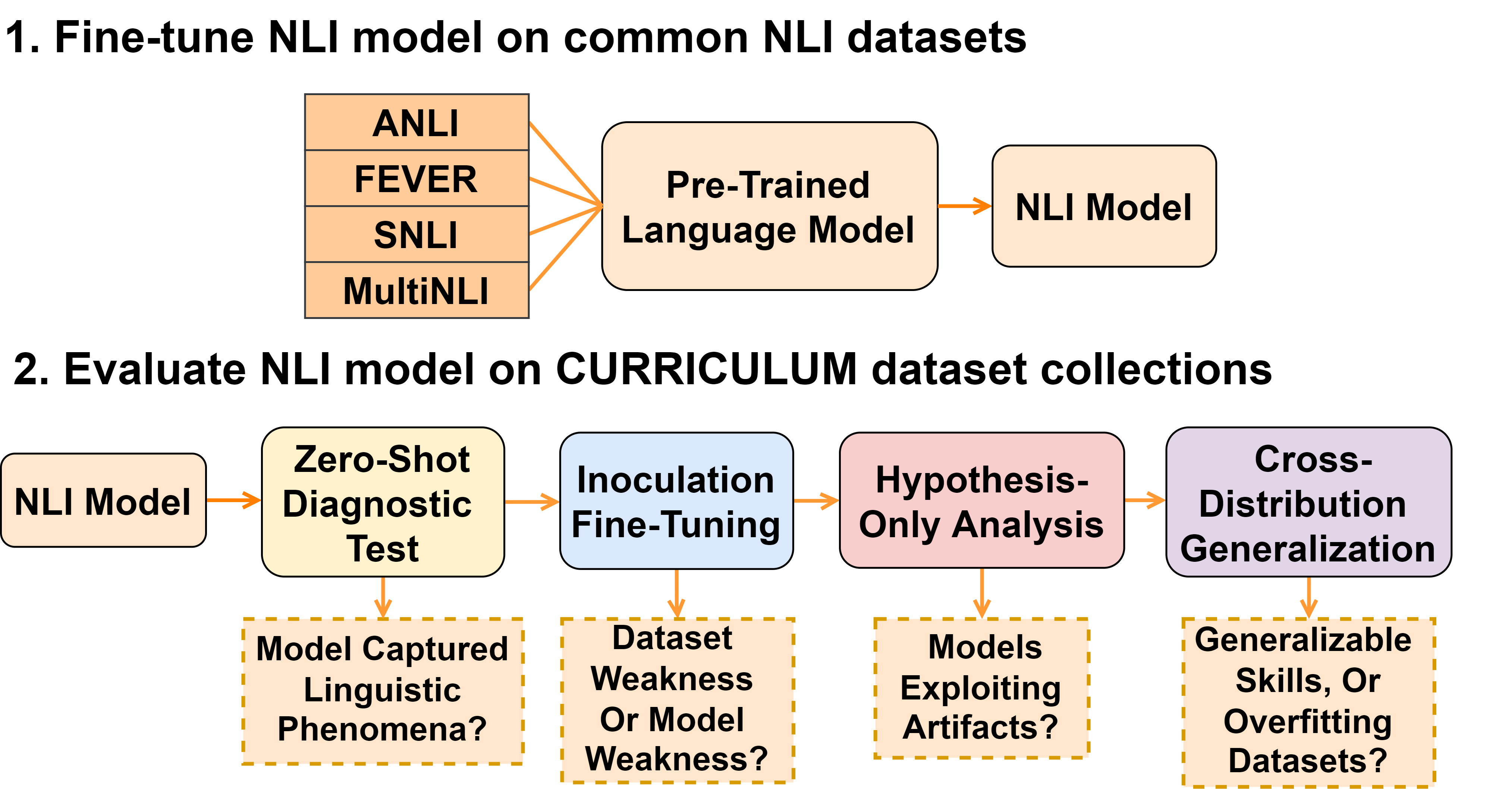}
    \caption{We propose a broad-coverage diagnostic benchmark for linguistic-phenomena-driven 
    evaluation. Our benchmark includes both a dataset collection and an evaluation procedure for evaluating model performance and diagnosing linguistic skills captured by a model. We evaluate models fine-tuned on large NLI datasets through four types of diagnostic tests: zero-shot, inoculation, hypothesis-only, and cross-distribution.} 
    \label{fig:curriculum}
\end{figure}

In this paper, we propose a new form of benchmark that serves as a diagnostic evaluation tool for analyzing model linguistic skills. We present \textsc{Curriculum} benchmark: a framework for diagnosing neural language models through broad-coverage linguistic phenomena. Our benchmark includes (1) a large-scale collection of natural language inference (NLI) datasets covering 36 linguistic phenomena and (2) an evaluation procedure for probing and evaluating how well a language model captures reasoning skills for distinct types of linguistic phenomena. Targeted linguistic phenomena in \textsc{Curriculum} range from fundamental properties like named entity and coreference to complex ones like commonsense and deductive reasoning. With the \textsc{Curriculum} benchmark, we aim to investigate the following research questions:
\begin{itemize}[leftmargin=*,nosep]
    \itemsep0em 
    \item \textbf{Q1}: Do language models trained on benchmark datasets have the ability to reason over a wide range of linguistic phenomena?
    \item \textbf{Q2}: Are linguistic phenomena missing from the training data recoverable through inoculation (i.e., continuing to train models on a small sample of examples) \cite{liu-etal-2019-inoculation}?
    \item \textbf{Q3}: Do language models learn a general reasoning skill of a phenomenon through inoculation?
\end{itemize}
To address the above questions, we empirically analyze NLI models trained on popular benchmark datasets
through a pipeline of evaluations that includes: a zero-shot diagnostic test, inoculation re-training, hypothesis-only sanity check, and cross cross-distribution generalization tests.

For \textbf{Q1}, we observe that models trained on benchmark datasets, including adversarial data, do not have the reasoning ability for a large set of linguistic phenomena. Our results show that training on more datasets can help the model learn more types of reasoning but does not help the model acquire complex reasoning skills such as deductive and commonsense reasoning. Our benchmark exposes multiple knowledge gaps in large NLI models regarding diverse linguistic phenomena, particularly in the categories of commonsense and comprehension. For \textbf{Q2}, our analysis provides empirical evidence that exposes the lack of recoverable linguistic phenomena in benchmark datasets and models' inability to learn certain linguistic phenomena. We also show that, on some phenomena, models may rely heavily on spurious dataset bias existing in the hypothesis to reach high accuracy. For \textbf{Q3}, Our experiments show that models can adapt between distributions with different difficulties only on 22.2\% of the phenomena such as Boolean, conditional, and comparative logic. In the majority (58.3 \%) of the phenomena, models fail to generalize when the difficulties of the train and test distributions are different, for example, relational knowledge, puns, and contextual commonsense reasoning. A model's learning performance may not align with its generalization ability, suggesting the lack of a general reasoning skill.

Overall, our proposed benchmark systematically maps out a wide range of specific linguistic skills required for language understanding and inference. We envision linguistic-phenomena-based evaluation to be an integral component of general linguistic intelligence. We hope \textsc{Curriculum} can serve as a useful evaluation tool that can map out which aspects of the problem space remain challenging for existing systems and models.

\section{Related Work}

\paragraph{NLU Benchmarks}
In recent years, multiple large-scale benchmarks for evaluating models' general language understanding performance have been proposed. Similar to our benchmark's task format, SNLI \cite{bowman-etal-2015-large} and MultiNLI \cite{williams-etal-2018-broad} are the two common benchmarks for Natural Language Inference (NLI). GLUE and SuperGLUE are the two most popular benchmarks that aim to provide a straightforward comparison between task-agnostic transfer learning techniques. They cover various task formats, task domains, and training volumes, with datasets all collected from publicly available sources. The construction of our benchmark is similar in that we also collect publicly available datasets from peer-reviewed papers. Adversarial NLI (ANLI) is a new benchmark collected "via an iterative, adversarial human-and-model-in-the-loop procedure." \cite{nie-etal-2020-adversarial}. ANLI is shown to be a more difficult challenge than previous benchmarks. Different from these benchmarks, our work aims to map out and evaluate specific linguistic skills a model needs for language understanding.  

\begin{table*}[]
    \centering
    \setlength\extrarowheight{3pt}
    \small
    \scalebox{0.87}{
    \begin{tabular}{lll} \toprule
         \textbf{Category} & \textbf{Description} & \textbf{Phenomena} \\ \midrule
         \multirow{3}{*}{\textbf{Lexical}} &  Testing a model's Word-level reasoning & Lexical Entailment (lex-ent), Named Entity (ner) \\ 
         & skill on lexical semantic,  direct, transitive, & Hypernymy (hyper), Hyponymy (hypo) \\  
         & and compositional lexical relationships. & Veridicality \& Transitivity (transit) \\\hline
         
         \multirow{2}{*}{\textbf{Syntactic}} & Testing a model's reasoning skill on & Syntactic Alternation (syn-alt),  VerbNet (vbn) \\ 
         & syntactic structure and compositionality. & Syntactic Variation (syn-var), VerbCorner (vbc) \\  \hline
         
         \multirow{3}{*}{\textbf{Semantic}} & Testing a model's reasoning skill on sentence-level reasoning & Sentiment (senti), Relational Knowledge (kg-rel), \\ 
         & involving diverse semantic properties: entity relations, & Puns (puns), Semantic Proto Label (sprl)\\ 
         & context, events, subjectivity, and semantic proto roles. & Context Alignment (ctx-align), Coreference (coref) \\ \hline
         
         \multirow{3}{*}{\textbf{Logical}}  & Testing a model's reasoning skill on logical operations:& Boolean (bool), Counting (count), Conditional (cond) \\
         & propositional structure, quantification, negation, & Comparative (comp), Negation (negat) \\ 
         & and monotonicity reasoning. & Monotonicity (monot), Quantifier (quant) \\\hline
         
          \multirow{3}{*}{\textbf{Analytical}}  & Testing a model's knowledge exploitation ability: drawing & Entailment Tree (ent-tree) \\ & accurate conclusions based on domain-specific knowledge, & Analytical  Reasoning (analytic) \\ 
          &  symbolic knowledge, and interpretable reasoning steps. & \\ \hline
         
          \multirow{2}{*}{\textbf{Commonsense}}  & Testing a model's reasoning skill on commonsense knowledge & Physical (physic), Social (social), HellaSwag (swag) \\
          & independent of cultural and educational background. & Contextual Commonsense Reasoning (cosmo) \\ \hline 
         
          \multirow{3}{*}{\textbf{Comprehension}}  & Testing a model's reasoning skill on complex reading & Event Semantics (ester), Discrete Reasoning (drop) \\ 
          & comprehension and inference, covering aspects of & Deductive Reasoning (logi) \\ 
          & semantic, context, logic, and numerical & Long Contextual Reasoning (control) \\ \hline
         
          \multirow{3}{*}{\textbf{Special}} & Testing a model's everyday reasoning skill. Including & Spatial Reasoning (spat), Temporal Reasoning (temp) \\
          & non-monotonic reasoning about valid but defeasible hypothesis & Defeasible Reasoning (defeas) \\
          & from hypothetical context and spatial-temporal reasoning. & Counterfactual Reasoning (counter) \\ \bottomrule
    \end{tabular}}
    \caption{This table lists the eight categories of linguistic phenomena covered by our dataset collection. We provide a brief introduction for each category describing the types of linguistic skills they intend to evaluate. We also list the dataset names and abbreviations each category contains.}
    \label{tab:categories}
\end{table*}

\paragraph{Fine-grained NLU Evaluation}
On top of large-scale benchmarks, there are several works \cite{joshi-etal-2020-taxinli, lognli} contributing to the fine-grained analysis of model performance. They collect data examples from existing benchmarks by attaching taxonomic category labels to each data. Or, they build semi-synthetic data allowing analysis on 17 reasoning dimensions. Our data collection and categorization concepts are similar to them. However, our work covers more linguistic phenomena that are difficult but important such as commonsense and non-monotonic reasoning. 

\paragraph{Challenge Datasets for NLU}
Many challenge datasets have been developed to evaluate models on specific linguistic skills for understanding. These datasets are in different formats such as NLI, Question Answering (QA), and Reading Comprehension (RC). They target a large set of skills including monotonicity \cite{yanaka-etal-2019-neural}, deductive logic \cite{liu2020logiqa}, event semantics \cite{han2021ester}, physical and social commonsense \cite{sap-etal-2019-social, bisk2019piqa}, defeasible reasoning \cite{rudinger-etal-2020-thinking}, and more. Our work brings together a set of challenge datasets to build a benchmark covering a large set of specific linguistic skills. We also merge different evaluation methods proposed by these works into a complete evaluation pipeline for our benchmark. 

\paragraph{Probing Linguistic Knowledge} 
Several works have found evidence that pre-trained models' representations encode knowledge about linguistic phenomena. \citet{tenney2018what} probe contextual representations from four pre-trained language models through the edge-probing method across tasks ranging from syntactic and semantic phenomena. They find that pre-trained models encode rich information on syntactic phenomena but only weakly encode information on semantic tasks compared to non-contextual baselines. \citet{abs-2112-01753}'s linguistic-information-probing framework extends the edge-probing study by focusing on different semantic phenomena that are important for logical inference in natural language. Their results show that pre-trained contextual embeddings encode more linguistic information on simple semantic phenomena than complex phenomena. Our work is partly motivated by this line of work in which our evaluation is based on the fact that pre-trained models can capture specific linguistic skills from learning. 

Other work investigates if models use specific linguistic skills to solve a downstream task. The DNC benchmark \cite{poliak-etal-2018-collecting} provides a collection of datasets for analyzing if models use distinct linguistic phenomena to conduct natural language inference. Several tasks in our benchmark come directly from this collection. However, our benchmark covers a wider range of linguistic phenomena from more categories than DNC. In particular, our benchmark contains semantic phenomena and includes phenomena from fundamental linguistic properties to complex reasoning types. In addition, our benchmark includes a systematic evaluation methodology that allows a more in-depth analysis of model behavior. 
\section{The \textsc{Curriculum} Benchmark}

\subsection{A New Form of Benchmark}
Recently, \citet{raji2021ai} suggested that good benchmark construction should focus on mapping out a specific set of linguistic skills required for language understanding. They recommend a future benchmark should provide interpretation on how systems work and how they fail on particular aspects of a problem space. Following this suggestion, we propose a new form of benchmark: linguistic-phenomena-driven evaluation. Our main objective is to reformulate the benchmark not simply to be a scoreboard for SOTA model contest but rather as a real measurement and standardization tool for (1) analyzing model performance, (2) exposing model and dataset weakness and (3) providing insights for future research directions.   

The curriculum benchmark aims to map out a specific set of linguistic skills required for language understanding. Our benchmark will serve as a diagnostic framework for linguistic-phenomena-driven probing and evaluation. The targeted linguistic skills should range from fundamental linguistic properties to complex reasoning types. Our linguistic phenomena selection is motivated by three benchmarks: GLUE Diagnostic, Rainbow, and DNC. In addition, we include many more phenomena focusing on complex reasoning types such as deductive logic and analytical thinking. Our finalized benchmark covers eight categories of linguistic phenomena. Each linguistic phenomenon is considered one task, and one should train, evaluate, and analyze models on each phenomenon individually. We briefly describe the types of reasoning skill each category focus on in Table \ref{tab:categories}. Appendix \ref{section:phenomena_ref} and \ref{section:phenomena_det} shows a list of references and dataset details for the train and test datasets used for each linguistic phenomenon.

\subsection{Dataset}
We collect many challenge NLI or NLU datasets and filter them individually with the following criteria: (1) We focus on datasets that evaluate a specific or a set of specific linguistic phenomena. (2) We focus on English monolingual datasets that are institutional and publicly available. (3) We exclude tasks that require domain-specific knowledge that we would not expect a model to learn through pre-training, such as medical knowledge. We finalize our selection with 36 datasets. Figure \ref{tab:categories} shows a detailed ontology of our selected linguistic phenomena and their abbreviations. Our motivation for dataset selection is mainly based on the linguistic phenomena categories that we aim to cover which will range from a simple to complex setting.

\subsection{Unified Task Format}
We unified the task formats into a single linguistic task, Natural Language Inference (NLI). NLI is a task for Natural Language Understanding. The task requires a model to classify the logical relationship between premise and a hypothesis. This logical relationship can either be Entailment (premise is true implies the hypothesis is absolutely true), Contradiction (premise is true implies the hypothesis is absolutely false), and Neutral (one cannot determine if the hypothesis is true or false based on the premise) \cite{2013Dagan}. We select NLI as the universal task format because NLI often serves as a general evaluation method for models on different downstream tasks. A model would need to handle nearly the full complexity of natural language understanding in order to solve the NLI task \cite{poliak-etal-2018-hypothesis}. Our benchmark contains two types of NLI problems: (1) the 3-way NLI with \texttt{Entailment}, \texttt{Contradiction}, and \texttt{Neutral}; (2) the 2-way NLI with \texttt{Entailed} and \texttt{Not-Entailed}. Each example has a premise and a hypothesis with 2-way or 3-way labels. 

\subsection{Automatic Recast}
To convert non-NLI datasets into the NLI task format, we follow the dataset recast procedure \cite{poliak-etal-2018-hypothesis}: automatically convert from non-NLI datasets with minimum human intervention. We design algorithmic ways to generate sentence pairs from the input text and convert the original labels into the NLI labels. Question Answering (QA) and Reading Comprehension (RC) are the two major tasks we need to convert. To convert datasets into NLI format, we follow the standard procedure \cite{Khot_Sabharwal_Clark_2018}. In QA datasets, if choices are given as declarative statements, we consider them as hypotheses and the question context as the premise. If choices are given as phrases answering the question, we concatenate the context and question to form a premise and consider the answers as hypotheses. Several datasets are tasks with free-response problems, and an answer can only be converted to an entailed hypothesis. To generate non-entailed hypotheses, we use several techniques during recasting. We show more details on our conversion techniques in Appendix \ref{section:data_recasting}. As a sanity check on our resulting datasets, we empirically find low performance on standard partial-input baselines \cite{poliak-etal-2018-hypothesis}, suggesting that our conversion yields data of high quality.

\begin{table}[t!]
    \centering
    \small
    \setlength\extrarowheight{2pt}
    \scalebox{0.9}{
    \begin{tabular}{lr lr lr} \toprule
         $\mathcal{P}$ &  $\mathrm{I}_v$ & $\mathcal{P}$ &  $\mathrm{I}_v$ & $\mathcal{P}$ &  $\mathrm{I}_v$   \\ \midrule
         lex-ent & 0.31 & transit & 0.41 & hyper & -0.99 \\ \hline
         hypo & -0.10 & ner & 0.19 & vbn & 0.55 \\ \hline 
         vbc & -0.40 & syn-alt & 0.10 & syn-var & 0.11 \\  \hline
         bool & 1.12 & cond & 1.13 & cont & 0.75 \\ \hline
         comp & 0.98 & negat & 1.13 & quant & 0.78 \\ \hline
         monot & -1.57 & kg-rel & 0.05 & coref & -0.38 \\ \hline
         senti & 0.42 & ctx-align & -0.79 & puns & 0.14 \\ \hline
         sprl & -0.11 & ent-tree & 0.50 & analytic & 0.00 \\ \hline
         temp & 0.10 & spat & 0.49 & counter & 0.47 \\ \hline 
         defeas & -0.39 & social & -0.40 & physic & -0.17 \\ \hline 
         swag & -0.66 & cosmo & -0.57 & drop & 0.19 \\ \hline 
         ester & -0.10 & logi & -0.71 & control & -0.07 \\
         \bottomrule
    \end{tabular}}
    \caption{Dataset difficulty measured by the amount of usable information ($\mathrm{I}_v$) from input data instances. The lower $\mathrm{I}_v$ is the more difficulty a dataset will be for the model. $\mathcal{P}$ here are the abbreviations of linguistic phenomena listed in Table \ref{tab:categories}}
    \label{tab:v_info}
\end{table}

\subsection{Dataset Difficulty}
To enhance our benchmark to provide more information on each dataset for in-depth evaluation and analysis, we provide each phenomenon a difficulty level. We use the predictive $\mathcal{V}$-information  \cite{ethayarajh2021informationtheoretic} as a measurement for dataset difficulty. The $\mathcal{V}$-information can measure how much information an input variable X can provide about Y when constrained to functions $\mathcal{V}$. Intuitively, more usable infromation X can provide, the easier a dataset is for the functions $\mathcal{V}$. Formally, let $\varnothing$ denote a null input that provides
no information about Y and $\mathcal{V}$ as a predictive family, we can compute the $\mathcal{V}$-information $\mathrm{I}_v(\mathrm{X} \rightarrow \mathrm{Y})$ as follows:
\begin{align*}
    \mathrm{H}_v(Y) &= \inf_{f\in\mathcal{V}} \expect[-\log f[\varnothing](\mathrm{Y})] \\
    \mathrm{H}_v(Y | X) &= \inf_{f\in\mathcal{V}} \expect[-\log f[\mathrm{X}](\mathrm{Y})] \\
    \mathrm{I}_v(\mathrm{X} \rightarrow \mathrm{Y}) &= \mathrm{H}_v(Y) - \mathrm{H}_v(Y | X)
\end{align*}
where $\mathrm{X}, \mathrm{Y}$ denote random variables with sample spaces $\mathcal{X}, \mathcal{Y}$. According to \citet{ethayarajh2021informationtheoretic}, $\varnothing$ can be an empty string here as $f[\varnothing]$ models the label entropy.
This framework can naturally adapt to the calculation of the point-wise $\mathcal{V}$-information ($\mathrm{PVI}$) where we measure the difficulty of each data example. Given a training dataset $\mathcal{D}_{train} = \{(x_i, y_i)\}^n_{i=1}$
, and the predictive family $\mathcal{V}$, the $\mathrm{PVI}$ of a data instance $(x,y) \in \mathcal{D}_{train}$ is computed as:
\begin{align*}
    \mathrm{PVI}(x \rightarrow y) &= -\log_2 f[\varnothing](y) + \log_2 f'[x](y),
\end{align*} 
where $\varnothing$ is an empty string (null input) and $\{f,f'\} \subseteq \mathcal{V}$. $f'$ and $f$ are models fine-tuned from $\mathcal{D}_{train}$ and $\{ (\varnothing, y_i) | (x_i, y_i) \in \mathcal{D}_{train} \}$ respectively.
The $\mathcal{V}$-information framework can also serve as a difficulty measurement for datasets and can be computed explicitly by averaging over $\mathrm{PVI}$:
\begin{align*}
    \mathrm{I}_v(\mathrm{X} \rightarrow \mathrm{Y}) &= \frac{1}{n} \sum_i \mathrm{PVI}(x_i \rightarrow y_i)
\end{align*}
As Table \ref{tab:v_info} shows, the difficulty level ranges from negative to positive. The higher the $\mathcal{V}$-information is, the easier a dataset is for the model. 

\paragraph{Dataset Controlled Split}
For our model evaluation pipeline, we are interested in verifying model's ability to learn a generalizable reasoning skill on linguistic phenomena. In particular, we want to check if a model can generalize when its training and testing data distributions have different measurement of difficulty. Thus, we need to conduct controlled split on datasets based on the point-wise difficulty, i.e. the point-wise $\mathcal{V}$-information of their data examples. We first calculate the $\mathrm{PVI}(x \rightarrow y)$ for each phenomenon dataset, then we split each dataset into two portions: simple and hard, based on the calculation of each example's $\mathrm{PVI}$.

\begin{table}[t!]
    \centering
    \footnotesize
    \scalebox{0.8}{
    \begin{tabular}{llll} \toprule
         Name & Model & Train/Test & Accuracy \\ \midrule
         \multirow{2}{*}{roberta-mnli} & RoBERTa & \multirow{2}{*}{MNLI/MNLI} & \multirow{2}{*}{90.2\%}  \\ 
         & \cite{liu2019roberta} & & \\ \hline
         \multirow{2}{*}{bart-mnli}  & BART & \multirow{2}{*}{MNLI/MNLI} & \multirow{2}{*}{89.9 \%} \\ 
         & \cite{lewis-etal-2020-bart} & & \\ \hline
         \multirow{3}{*}{roberta-anli-mix} & \multirow{3}{*}{RoBERTa} & SNLI, MNLI, & \multirow{3}{*}{53.7 \%}  \\
         &  &  FEVER, ANLI/ &  \\ 
         &  & ANLI & \\ \hline
         \multirow{3}{*}{xlnet-anli-mix}  & \multirow{2}{*}{XLNet} & SNLI, MNLI & \multirow{3}{*}{55.1 \%} \\ 
         &  & FEVER, ANLI/ & \\
         &  \cite{Xlnet} & ANLI & \\ \bottomrule
    \end{tabular}}
    \caption{Details on models used in our experiments. All four models are large models and publicly available.}
    \label{tab:models}
\end{table}

\section{Evaluation Methodology}
We define an evaluation process for the \textsc{Curriculum} benchmark that aims to bring different types of evaluation and diagnosing methods used by previous challenge NLI datasets. Following \citet{raji2021ai}'s suggestion, we want our evaluation process to both to analyze the model output in detail and explore which aspects of the inference problem space remain challenging to current models.

\paragraph{Zero-shot Diagnostic Test}
This test is motivated by the diagnostic test in GLUE. We focus on providing fine-grained analysis of zero-shot system performance on a broad range of linguistic phenomena. We follow the GLUE diagnostic dataset and use the Matthews Correlation Coefficient (MCC) \cite{10.1371/journal.pone.0041882} as the evaluation metric. MCC computes the correlation coefficient of the predicted labels and the true labels. The correlation coefficient value is between -1 and +1. A coefficient of +1 indicates a perfect prediction. A 0 indicates average random prediction A -1 indicates the classifier always miss-classifies. MCC is perfectly symmetric, so it can be used even if the dataset has classes with different sizes.

\paragraph{Inoculation by Fine-tuning}
We use inoculation \cite{liu-etal-2019-inoculation} to further analyze model failures on target linguistic phenomena. This method fine-tunes the model on a down-sampled training section of a phenomenon dataset (inoculation). One can interpret inoculation performance in two ways:
\begin{enumerate} [leftmargin=*,nosep]
    \item Good performance: the original training set of the model, prior to inoculation, did not sufficiently cover the target phenomenon, but it is recoverable through through additional training on a small sample of data.
    \item Poor performance: there exists a model weakness to handle the target phenomenon.
\end{enumerate}

\paragraph{Hypothesis-only Bias Analysis}
We conduct analysis on hypothesis-only bias as (1) a sanity check for our converted datasets and also and (2) a verification on whether model's good performance is from leveraging artifacts in the hypotheses. We train a hypothesis-only baseline \cite{poliak-etal-2018-hypothesis} for each phenomenon and compare their performance against the best models from the inoculation experiment. We want to ensure that models' improved performance after inoculation is due to their ability to reason about a hypothesis and the given context together. If the hypothesis-only baseline shows good performance, we interpret this as a sign that the datasets contain artifact. If the baseline shows poor performance, it gives evidence that the model is not taking short-cuts.

\paragraph{Cross-Distribution Generalization}
We conduct the cross-distribution generalization test \citep{rozen-etal-2019-diversify} to verify if the model learns a general reasoning skill from inoculation. The good inoculation performance does not ensure that the model's learned skill is generalizable. The model can likely over-fit the dataset distribution by adopting superficial cues. We evaluate the model's generalization ability by training and testing the model on distributions yielding different difficulty levels within the same dataset. For example, we train the model on the simple part of the dataset (data with high $\mathcal{V}$-information) and test it on the hard part (data with low $\mathcal{V}$-information).

\begin{figure*}[t!]
     \centering
     \begin{subfigure}[b]{0.9\textwidth}
         \centering
         \includegraphics[width=\textwidth]{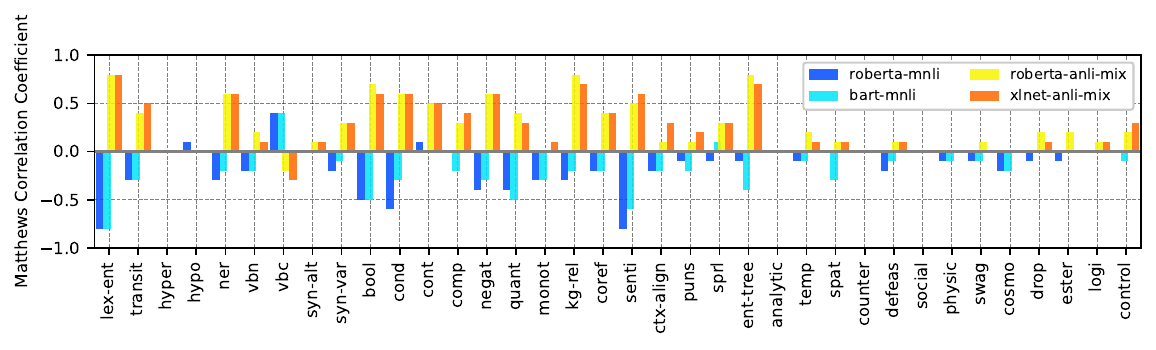}
         \caption{\small Zero-shot system performance on the \textsc{Curriculum} benchmark.}
         \label{fig:zero-shot}
     \end{subfigure}
     \vfill
     \begin{subfigure}[b]{0.9\textwidth}
         \centering
         \includegraphics[width=\textwidth]{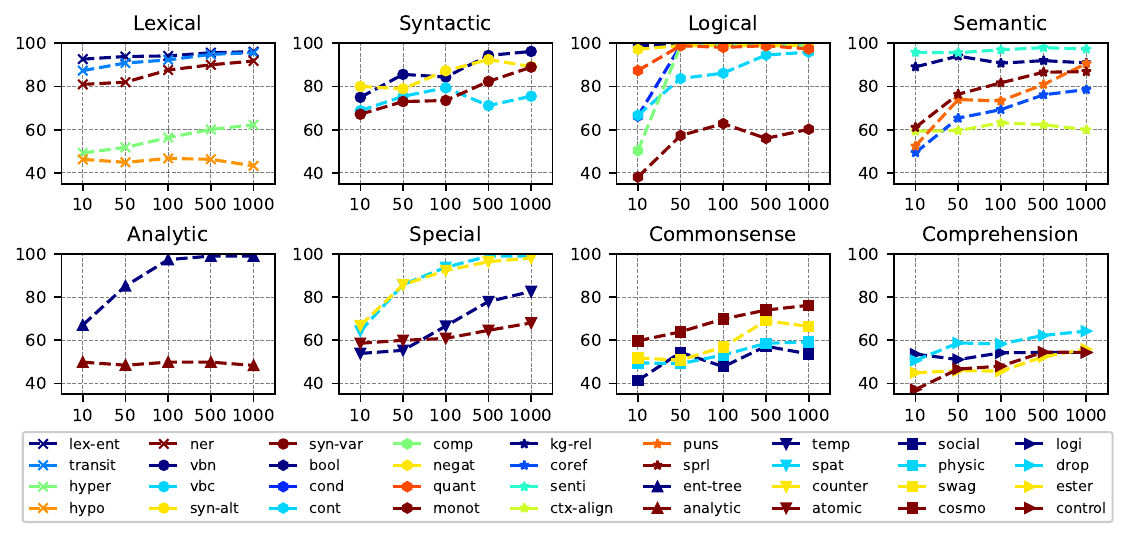}
         \includegraphics[width=\textwidth]{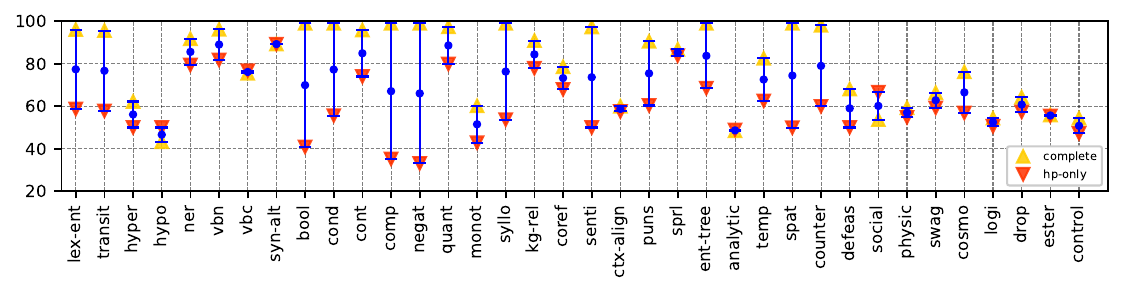}
         \caption{\small Inoculation by fine-tuning vs. hypothesis-only analysis. The X-axis of the top plot represents training examples per label. Both plots' Y-axis show the accuracy. Models used in these two experiments are both the roberta-anli-mix model, introduced in Section 4.1.}
         \label{fig:inoculate}
     \end{subfigure}
\end{figure*}

\subsection{Experiment Setup}
For the zero-shot test, we test a model on each test set without additional fine-tuning. We select NLI models with top performance on NLI benchmarks MNLI and ANLI. We list these models in Table \ref{tab:models}. We are interested in evaluating models with both the single-encoder and the text2text architecture. All models are publicly available from Huggingface \cite{arxiv.1910.03771}. For inoculation, we fine-tune models on training examples with a size ranging from 10 to 1000 examples per label. For the cross-distribution generalization test, we first create variant data distributions for train and test sets using the $\mathcal{V}$-information-based dataset split method from Section 3.5. We split each dataset into two portions (simple and hard) according to the point-wise $\mathcal{V}$ information. Next, we either train and test the model on the same difficulty distribution or train it on one portion and test it on a different portion. In the inoculation, hypothesis-only, and generalization experiments, we all use roberta-anli-mix as our NLI model because its training set covers all the major NLI training datasets: SNLI, MNLI, FEVER \cite{thorne-etal-2018-fever}, and ANLI. We use accuracy as our evaluation metric for all these three experiments. For all the experiments excluding zero-shot test, we run several turns and select the best performance for analysis. 
\section{Empirical Analysis}
\label{sec:analysis}

\begin{figure*}[t!]
    \centering
    \includegraphics[width=0.9\textwidth]{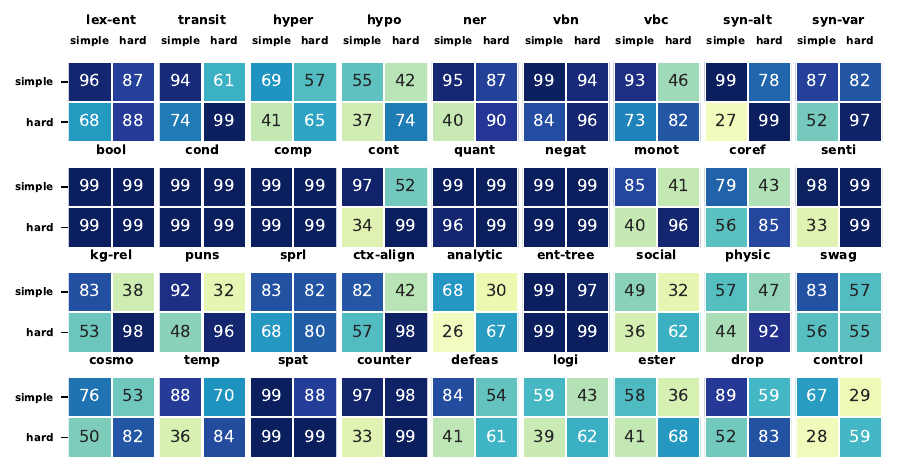}
    \caption{Generalization between controlled dataset splits. Here each heat-map shows the generalization performance of the model fine-tuned and evaluated on different distributions within each linguistic phenomenon.}
    \label{fig:generalize}
\end{figure*}

\subsection{Zero-shot Linguistic Phenomena Diagnose}
First, we report the results on zero-shot diagnostic evaluation for each baseline model. From Figure \ref{fig:zero-shot}, we observe that both single-encoder and text2text models trained on MultiNLI show a negative correlation in the majority of linguistic phenomena. Meanwhile, anli-mix models (roberta-anli-mix, xlnet-anli-mix) are positively correlated on most (77.8 \%) of the phenomena and they show high correlation ($> 0.50$) on 27.8 \% of the phenomena. On average, models trained on the large dataset mixture show better performance than models trained on MultiNLI alone, suggesting that training on more datasets help models capture more types of linguistic phenomena. However, most of the phenomena captured by the anli-mix models are easier to learn (higher $\mathcal{V}$ information). On harder phenomena, models did not benefit from the training dataset mixture. For instance, both the anli-mix models have a low correlation on deductive and analytical reasoning. Overall, we find that NLI datasets from common benchmarks lack examples of a diverse set of reasoning skills.



\subsection{Inoculation}
Based on Figure \ref{fig:inoculate}, the model can reach high accuracy on about 64 \% of the phenomena as the training examples accumulate. Most of these phenomena have higher $\mathcal{V}$ information ($> 0.0$) that should relatively be easier to learn. We are surprised that for some hard phenomena ($\leq 0.0$) such as commonsense contextual reasoning (cosmo, -0.67), the model's performance improved after inoculation. The improvement shows an gap in the original training data mixture. On 25 \% of the phenomena, the model's performance did not improve significantly after inoculation, meaning that it fails to learn the reasoning skills for these phenomena. Most of these phenomena are difficult, with a low $\mathcal{V}$ information, such as monotonicity(mono) and deductive (logi) reasoning. The accuracy is consistently low when training examples accumulate. 

We also observe that model struggles to learn phenomena that require complex reasoning, such as phenomena from the comprehension category. This trends show inherent weaknesses in the model or its training strategy that cause its failure to learn complex and hard phenomena. Overall, results from this experiment, combined with the zero-shot evaluation, suggest that many linguistic phenomena are missing from different large-scale NLI datasets but are recoverable through additional training examples. However, the model fails to learn the skills for hard and complex phenomena. In summary, our diagnostic study through inoculation exposes a diverse set of dataset and model weaknesses. 

\subsection{Hypothesis-only Bias}
To determine if models can leverage spurious artifacts in the hypotheses of each phenomenon, we compare full models to hypothesis-only baselines. From Figure \ref{fig:inoculate}, we observe that hypothesis-only baseline performs poorly on a majority of the phenomena. This indicates that our benchmark generally requires the model to learn an inference process between contexts and hypotheses for good performance. We observe that on 30.6\% of the phenomena, the full-model can reach a high accuracy while the baseline has low accuracy, suggesting the model can learn the phenomenon without relying on hypothesis artifacts. On 36 \% of the phenomena, the model does not show a significant performance gain compared to the baseline. Most of these are complex reasoning phenomena like deductive and analytical reasoning. The result validates that the model struggles more with complex linguistic phenomena. On 33.3 \% of the phenomena, both the full-model and the baseline achieve high accuracy showing the possibility that the model exploits artifacts from the hypothesis to reach high accuracy. 

Also, note that the hypothesis-only baseline performs better for some tasks than the fine-tuned model, which can be interpreted in two ways. When both the baseline and fine-tuned model achieve high accuracy (vbc, syn-alt), higher accuracy on baseline indicates that the hypothesis-only bias is pretty strong in the dataset. When the intervention from the premise is removed (hypothesis-only input), the models can easily exploit the bias to achieve higher accuracy. In contrast, when both the baseline and fine-tuned model achieve low accuracy (hypo, analytic, social, ester), higher accuracy on baseline indicates that the task is very difficult for a model to master successfully. Low baseline accuracy means that the dataset does not contain much bias, so a model must learn the correct reasoning to perform well. However, the fine-tuned model has even worse performance than the baseline, meaning that it fails to learn the skill required for these tasks. Our main finding here is that good performance on a linguistic phenomenon dataset does not mean the model captured the associated phenomena. The model can learn short-cuts through hypothesis-only bias and artifacts.



\subsection{Generalization}
As Figure \ref{fig:generalize} show, the model can adapt between different distributions only on 22.2 \% of the phenomena. The model achieves high accuracy consistently for all four categories in the generalization matrix suggesting the learned skills are generalizable. On 58.3 \% phenomena, models can not generalize between different difficulty distributions. They show higher accuracy when trained and tested on the same distribution but low accuracy when the test distribution shifted. For example, on relational knowledge reasoning (kg-rel), the model achieves 83\% for simple $\rightarrow$ simple and 98 \% for hard $\rightarrow$ hard. Nevertheless, the performance drops to 53 \% for hard $\rightarrow$ simple and 38 \% for simple $\rightarrow$ hard. 

Notice that model's good performance on inoculation does not align with its generalization ability. For example, the model reaches 90.9 \% accuracy on kg-rel, but its generalization performance is poor. This behavior highlights a model weakness: can over-fit to a particular distribution but fail to learn a general reasoning skill for the target phenomenon. We observe an interesting behavior that models struggle to generalize from hard to simple distribution on about 14 \% of the phenomena while showing good generalization from simple to hard distribution. We think the possible reason is that the hard distribution contains data with relatively low $\mathcal{V}$ information. A low amount of usable information makes it hard for the model to learn the phenomena sufficiently for generalization.

\section{Conclusion and Future Work}
In this paper, we introduce a new form of benchmark that can serve as an effective tool for evaluating and analyzing model outcomes. We propose a linguistic-phenomena-driven benchmark that aims to diagnose neural language models to discover types of linguistic skills that remain challenging to models. We compiled a dataset collection covering 36 types of linguistic phenomena ranging from fundamental linguistic properties to complex reasoning skills. In addition, we define an evaluation procedure that can provide an in-depth analysis of model and dataset weaknesses. Using our benchmark, we comprehensively study how well language models capture specific linguistic skills essential for understanding. Our major findings include:
\begin{itemize}[leftmargin=*,nosep]
    \item Models trained on benchmark NLI datasets fail to reason over a diverse set of linguistic phenomena.
    \item Good inoculation performance on some phenomena results from the model leveraging superficial artifacts in the hypothesis.
    \item The model tends to over-fit the dataset distribution without learning a general reasoning skill on a majority of phenomena.
\end{itemize}
Overall, our benchmark effectively evaluates a model on specific linguistic skills and exposes a list of model and training data weaknesses. We hope that our benchmark and empirical findings can encourage the community to rethink dataset construction and model architecture design. In particular, we hope to encourage the development of new datasets that cover richer types of linguistic phenomena and language models that can learn generalizable linguistic skills. For future work, we plan to add more datasets to cover more phenomena such as psycho-linguistics \cite{laverghetta-jr-etal-2021-transformer}. We envision our benchmark to be dynamic, meaning that a dataset with higher quality and difficulty for a phenomenon should replace the current ones in the future. For example, the StepGame benchmark \cite{stepgame} provides better data for spatial reasoning, which can replace the current spatial reasoning dataset. We also plan to explore new learning methods to help models overcome the weakness of learning non-generalizable skills, such as calibration through symbolic loss functions.



\section*{Acknowledgment}
We thank the anonymous reviewers for their thoughtful and
constructive comments. We thank Kyle Richardson from AI2 for his insights and suggestions on improving our camera-ready version. Thanks also to our advisors Laurence S. Moss and Michael Wollowski for their feedback on earlier drafts of this work. Special thanks to the Machine Learning for Language Group at NYU for their wonderful NLP toolkit, JIANT \cite{phang2020jiant}.

\bibliography{anthology, misc}
\bibliographystyle{acl_natbib}

\clearpage
\appendix
\clearpage
\onecolumn

\vspace{0.5cm}

\section{Linguistic Phenomena in \textsc{Curriculum}}
\label{section:phenomena_ref}

\begin{table*}[h!]
    \centering
    \small
    \scalebox{0.9}{
    \begin{tabular}{p{14em} p{17em} p{20em}}
         \toprule
         \textbf{Phenomena} &  \textbf{Train Reference} & \textbf{Test Reference}\\
         \midrule
         \multicolumn{3}{c}{\textbf{Lexical Phenomena}} \\
         \midrule
         
         Lexical Entailment 
         & \citealt{schmitt-schutze-2021-language}
         & \citealt {schmitt-schutze-2021-language, glockner-etal-2018-breaking} \\
         
         Hypernymy
         & \citealt{richardson-sabharwal-2020-qa}
         & \citealt{richardson-sabharwal-2020-qa} \\
         
         Hyponymy
         & \citealt{richardson-sabharwal-2020-qa}
         & \citealt{richardson-sabharwal-2020-qa} \\
         
         Named Entity
         & \citealt{poliak-etal-2018-collecting}
         & \citealt{poliak-etal-2018-collecting} \\ 
         
         Veridicality and Transitivity    
         & \citealt{poliak-etal-2018-collecting, yanaka-etal-2021-exploring}
         & \citealt{poliak-etal-2018-collecting, yanaka-etal-2021-exploring} \\ \midrule
         
         \multicolumn{3}{c}{\textbf{Syntactic Phenomena}} \\
         \midrule
         
         VerbNet
         & \citealt{poliak-etal-2018-collecting}
         & \citealt{poliak-etal-2018-collecting} \\
         
         VerbCorner
         & \citealt{poliak-etal-2018-collecting}
         & \citealt{poliak-etal-2018-collecting} \\ 
         
         Syntactic Variation  
         & \citealt{dolan-brockett-2005-automatically}
         & \citealt{dolan-brockett-2005-automatically}  \\
         
         Syntactic Alternations   
         &  \citealt{kann-etal-2019-verb} 
         &  \citealt{kann-etal-2019-verb} \\ \midrule
         
         \multicolumn{3}{c}{\textbf{Semantic Phenomena}} \\
         \midrule
         
         \multirow{2}{*}{Coreference \& Anaphora}
         & \citealt{sakaguchi2019winogrande, wang2019glue} & \citealt{sakaguchi2019winogrande, wang2019glue} \\
         & \citealt{webster-etal-2018-mind} & \citealt{webster-etal-2018-mind} \\
         
         Sentiment
         & \citealt{poliak-etal-2018-collecting}
         & \citealt{poliak-etal-2018-collecting} \\
         
         Relational Knowledge
         &  \citealt{poliak-etal-2018-collecting} 
         &  \citealt{poliak-etal-2018-collecting} \\
         
         Puns
         &  \citealt{poliak-etal-2018-collecting} 
         &  \citealt{poliak-etal-2018-collecting} \\
         
         Semantic Proto Label 
         &  \citealt{white-etal-2017-inference} 
         &  \citealt{white-etal-2017-inference} \\
         
         Context Alignment
         &  \citealt{white-etal-2017-inference} 
         &  \citealt{white-etal-2017-inference, bigbench}\\ \midrule
         
         \multicolumn{3}{c}{\textbf{Logical Phenomena}} \\ \midrule
         
         Boolean
         &  \citealt{richardson2019probing} 
         &  \citealt{richardson2019probing} \\
         
         Conditional
         &  \citealt{richardson2019probing} 
         &  \citealt{richardson2019probing} \\
         
         Comparative
         &  \citealt{richardson2019probing} 
         &  \citealt{richardson2019probing}  \\
         
         Counting
         &  \citealt{richardson2019probing} 
         &  \citealt{richardson2019probing} \\
         
         Quantifier
         &  \citealt{richardson2019probing} 
         &  \citealt{richardson2019probing}  \\
         
         Negation 
         &  \citealt{richardson2019probing}  
         &  \citealt{richardson2019probing}  \\
         
         Monotonicity 
         & \citealt{yanaka-etal-2019-help}
         & \citealt{yanaka-etal-2019-neural, richardson2019probing} \\ \midrule
         
         \multicolumn{3}{c}{\textbf{Analytic Phenomena}} \\
         \midrule
         
         Entailment Tree         
         & \citealt{dalvi2021explaining}
         & \citealt{dalvi2021explaining} \\
         
         Analytical Reasoning  
         &  \citealt{zhong2021arlsat}
         &  \citealt{zhong2021arlsat}  \\ \midrule

         \multicolumn{3}{c}{\textbf{Commonsense Phenomena}} \\
         \midrule

         Physical 
         & \citealt{bisk2019piqa}
         & \citealt{bisk2019piqa} \\
         
         Social      
         & \citealt{sap-etal-2019-social}
         & \citealt{sap-etal-2019-social} \\
         
         HellaSwag
         & \citealt{atomic}
         & \citealt{atomic} \\
         
         Contextual Commonsense 
         & \citealt{huang-etal-2019-cosmos}
         & \citealt{huang-etal-2019-cosmos} \\  
         Reasoning & &  \\ \midrule
         
         \multicolumn{3}{c}{\textbf{Comprehension Phenomena}} \\
         \midrule
         
         Deductive Reasoning
         & \citealt{liu2020logiqa}
         & \citealt{liu2020logiqa} \\ 
         
         Contextual Reasoning
         & \citealt{aaaiLiuCL021}
         & \citealt{aaaiLiuCL021} \\
         
         Event Semantic Reasoning 
         & \citealt{han2021ester} 
         & \citealt{han2021ester} \\ 
         
         Discrete Reasoning
         & \citealt{dua-etal-2019-drop} 
         & \citealt{dua-etal-2019-drop}  \\

         \midrule
         \multicolumn{3}{c}{\textbf{Special Reasoning Phenomena}} \\
         \midrule
         
         Defeasible Reasoning 
         & \citealt{rudinger-etal-2020-thinking}
         & \citealt{rudinger-etal-2020-thinking} \\
            
         Temporal Reasoning 
         & \citealt{bAbI} 
         & \citealt{bAbI} \\
         
         Spatio Reasoning 
         & \citealt{bAbI} 
         & \citealt{bAbI} \\
         
         Counterfactual Reasoning
         & \citealt{patil-baths-2020-cnrl}
         & \citealt{patil-baths-2020-cnrl} \\ \bottomrule
    \end{tabular}}
    \caption{A detailed list of training datasets and test datasets used for each linguistic phenomenon in our benchmark.}
    \label{tab:curriculum_reference}
\end{table*}

\clearpage

\section{\textsc{Curriculum} Dataset Details in \textsc{Curriculum}}
\label{section:phenomena_det}

\begin{table*}[h!]
  \setlength{\tabcolsep}{2pt}
  \small
  \centering
  \begin{tabular}{llrrr}
    \toprule
    & \textbf{Name} & \textbf{$|$Train$|$}  & \textbf{$|$Dev$|$}  & \textbf{Original task} \\ 
    \midrule

    & Lexical Entailment & 6398 & 2964 & NLI \\  
    & Hypernymy & 20000 & 8500 & QA\\
    & Hyponymy  & 20000 & 8500 & QA\\
    & Named Entity & 50000 & 30000 & NLI \\
    & Veridicality and Transitivity & 20000 & 8788 & NLI \\ \midrule
    
    & VerbNet & 1398 & 160 & NLI  \\
    & VerbCorner & 110898 & 13894 & NLI \\
    & Syntactic Variation & 3668 & 408 & SC \\
    & Syntactic Alternations & 19990 & 8739 & SC  \\ \midrule

    & Coreference \& Anaphora & 12135 & 5799 & NLI/SC \\
    & Sentiment & 4800 & 600 & NLI \\
    & Relational Knowledge  & 21905 & 761 & NLI \\
    & Semantic Proto Label & 14038 & 1756 & NLI \\ 
    & Puns & 14038 & 1756 & NLI \\ 
    & Context Align & 14038 & 1756 & NLI \\  \midrule
    
    & Boolean & 3000 & 1000 & NLI  \\
    & Conditional  & 3000 & 1000 & NLI \\
    & Comparative  & 3000 & 1000 & NLI  \\
    & Counting & 3000 & 1000 & NLI \\
    & Quantifier & 3000 & 1000 & NLI \\
    & Negation QA & 3000 & 1000 & NLI \\
    & Monotonicity & 35891 & 5382 & NLI \\ \midrule
    
    & Entailment Tree & 1314 & 340 & TG \\
    & Analytical Reasoning & 3260 & 922 & SC \\ \midrule
    
    & Physical & 10000 & 1838 & QA \\   
    & Social & 6003 & 6003 & QA \\   
    & HellaSwag & 20000 & 8518 & QA \\   
    & Contextual Commonsense Reasoning & 9046 & 5452 & RC \\   \midrule
    
    & Deductive Reasoning & 14752 & 2604 & RC \\   
    & Contextual Reasoning & 6719 & 1604 & RC \\   
    & Event Semantics Reasoning & 2800 & 662 & RC \\   
    & Discrete Reasoning & 20000 & 13148 & RC \\  \midrule
    
    & Defeasible Reasoning & 39036 & 9860 & SC \\ 
    & Temporal Reasoning & 4248 & 1174 & NLI \\   
    & Spatial Reasoning & 10000 & 10000 & QA \\   
    & Counterfactual Reasoning & 6062 & 3364 & SC \\
    \bottomrule
  \end{tabular}
  \caption{Overview of all the linguistic phenomena datasets in our benchmark. QA is short for Question Answering. NLI is short for Natural Language Inference. SC is short for Sentence Classification. TG is short for Text Generation. RC is short for Reading Comprehension.}
  \label{tab:dataset_det}
\end{table*}

\clearpage
\onecolumn

\vspace{0.5cm}

\section{Data Recasting Details}
\label{section:data_recasting}
Here we provide more details on the major techniques we used to convert Question Answering (QA) and Reading Comprehension (RC) datasets into recast NLI datasets.

\subsection{Entity Swapping}

\begin{table}[h!]
    \centering
    \small
    \scalebox{0.85}{
    \begin{tabular}{l}
        \textbf{<Original>} \\
         \textbf{Context: } \texttt{...The Buccaneers tied it up with a 38-yard field goal} \\
         \texttt{by Connor Barth, ... The game's final points came} \\
         \texttt{when Mike  Williams of Tampa Bay caught a 5-yard pass...} \\
         \textbf{Q: }\texttt{Who caught the touchdown for the fewest yard?} \\ 
        \textbf{Answer: } \texttt{Mike Williams} \\ \hline 
         \textbf{<Recast>} \\
         \textbf{Premise: } \texttt{...The Buccaneers tied it up with a 38-yard field goal} \\
         \texttt{by Connor Barth, ... The game's final points came} \\
         \texttt{ when Mike  Williams of Tampa Bay caught a 5-yard pass...} \\
         \textbf{Hypothesis: }\texttt{Mike Williams caught the touchdown for the fewest yard} \\
         \textbf{Label: }Entailed \\
         \textbf{Hypothesis: }\texttt{Connor Barth caught the touchdown for the fewest yard} \\
         \textbf{Label: }Not-Entailed \\
    \end{tabular}}
    \caption{Example of converting an RC example from DROP \cite{dua-etal-2019-drop} to NLI format. The entailed hypothesis is a concatenation of question and answer. The non-entailed hypothesis is created by entity swapping on the entailed one ({\texttt{Mike Williams} $\rightarrow$ \texttt{Connor Barth}}).}
    \label{tab:recast1}
\end{table}

\subsection{Question/Answer Concatenation}
\begin{table}[h!]
    \centering
    \small
    \scalebox{0.85}{
    \begin{tabular}{l}
        \textbf{<Original>} \\
         \textbf{Context: } \texttt{The flash in the room that followed was proof of that assumption. The man grabbed his arm again.} \\
         \texttt{"Please let go of my arm." He requested, his voice low. "Look."} \\
         \textbf{Q: }\texttt{Why did the man grabbed his arm?} \\ 
         \textbf{Choice 1:} The man wanted to dance with him. \\
         \textbf{Choice 2:} \textbf{\textit{The man wanted to get his attention.}} \\
         \textbf{Choice 3:} The man wanted to pull him closer so he can cry on this shoulder. \\
         \textbf{Choice 4:} The man was angry with him and wanted to push him outside. \\
        \hline 
         \textbf{<Recast>} \\
         \textbf{Premise: } \texttt{The flash in the room that followed was proof of that assumption. The man grabbed his arm again.} \\
         \texttt{"Please let go of my arm." He requested, his voice low. "Look."} \\
         \textbf{Hypothesis: }\texttt{The man wanted to get his attention.} \\
         \textbf{Label: }Entailed \\
         \textbf{Hypothesis: }\texttt{The man wanted to dance with him.} \\
         \textbf{Label: }Not-Entailed \\
    \end{tabular}}
    \caption{Example of converting an QA example from Cosmos QA \cite{huang-etal-2019-cosmos} to NLI format. The entailed hypothesis is the correct answer from the given choices. The non-entailed hypothesis is one of the false answers, excluding the choice "None of the above choices".}
    \label{tab:recast2}
\end{table}


\section{Reproducibility}
\paragraph{Implementation.} Our model training and testing pipeline is modified from the JIANT toolkit. We mainly adapted several components on classes and functions involving task, dataset,
reprocessing, tokenization, model version control, and evaluation metrics. All our experiments are implemented with models publicly available from Huggingface Transformers \cite{wolf-etal-2020-transformers}\footnote{\url{https://github.com/huggingface/transformers}}.

\paragraph{Hyper-parameters} We mainly follow the practice in \cite{nie-etal-2020-adversarial}. For all the experiments excluding the zero-shot test in Section 5.1, we use a learning rate of $1e-5$ with a batch size of 8. We set the number of warmup updates to be 1000. We set the epoch number to be 3 and 5. We evaluate the model on $D_{dev}$ every 200 steps for the inoculation and generalization experiments, and 500 steps for the hypothesis-only experiment. We use the AdamW \cite{adamw} as our optimizer.

\paragraph{Infrastructure} All experiments are done with one single Geforce RTX 3090 (24GB). A single inoculation or generalization job finishes within 0.5 hours on average. A single hypothesis-only job finishes within 1-2 hours on average.

\paragraph{Number of Parameters.} RoBERTa-large model contains 355 million parameters. BART-large model contains 139 million parameters. BART-Large model contains 406 million parameters. XLNet-large model contains 340 million parameters.
\end{document}